\documentclass[conference]{IEEEtran}
\usepackage{times}

\usepackage[numbers]{natbib}
\usepackage{multicol}
\usepackage{graphicx}
\usepackage{algorithm,algpseudocode}
\usepackage{algorithmicx,algpseudocode}
\usepackage[font={small}]{caption}
\usepackage{makecell}
\usepackage{amsmath}
\usepackage{color}

\begin{document}

\title{Online User Assessment for Minimal Intervention During Task-Based Robotic Assistance}

\author{\IEEEauthorblockN{
  Aleksandra Kalinowska\IEEEauthorrefmark{1},
  Kathleen Fitzsimons\IEEEauthorrefmark{1},
  Julius Dewald\IEEEauthorrefmark{2}\IEEEauthorrefmark{3}\IEEEauthorrefmark{4}, and 
  Todd D Murphey\IEEEauthorrefmark{1}\IEEEauthorrefmark{2}}

 \IEEEauthorblockA{\IEEEauthorrefmark{1}Department of Mechanical Engineering, Northwestern University,
 Evanston, IL}

 \IEEEauthorblockA{\IEEEauthorrefmark{2}Physical Therapy and Human Movement Science, Northwestern University, Chicago, IL}
 
 \IEEEauthorblockA{\IEEEauthorrefmark{3}Physical Medicine and Rehabilitation, Northwestern University, Chicago, IL}
 \IEEEauthorblockA{\IEEEauthorrefmark{4}Biomedical Engineering, Northwestern University, Chicago, IL}}

\maketitle

\begin{abstract}
We propose a novel criterion for evaluating user input for human-robot interfaces for known tasks. We use the mode insertion gradient (MIG)---a tool from hybrid control theory---as a filtering criterion that instantaneously assesses the impact of user actions on a dynamic system over a time window into the future. As a result, the filter is permissive to many chosen strategies, minimally engaging, and skill-sensitive---qualities desired when evaluating human actions. Through a human study with 28 healthy volunteers, we show that the criterion exhibits a low, but significant, negative correlation between skill level, as estimated from task-specific measures in unassisted trials, and the rate of controller intervention during assistance. Moreover, a MIG-based filter can be utilized to create a shared control scheme for training or assistance. In the human study, we observe a substantial training effect when using a MIG-based filter to perform cart-pendulum inversion, particularly when comparing improvement via the RMS error measure. Using simulation of a controlled spring-loaded inverted pendulum (SLIP) as a test case, we observe that the MIG criterion could be used for assistance to guarantee either task completion or safety of a joint human-robot system, while maintaining the system's flexibility with respect to user-chosen strategies. 
\end{abstract}

\IEEEpeerreviewmaketitle

\section{Introduction}

Shared control algorithms have been developed for robotic assistance and robot-supported training in applications ranging from assisted vehicle navigation~\cite{autovehicle2014} and surgical robotics~\cite{surgeryrobot2004,RSS_surgery2014} to  brain-computer interface manipulation~\cite{teleBCI2017} and exoskeleton-assisted gait~\cite{RSS_exo2013, exoMPC2011}. The aims and safety requirements of these systems vary greatly, but one challenge is often the same---how do we allocate control between the user and the machine? 

A factor to consider is user preference. In~\cite{RSS_userpref2017}, machine learning techniques were used to model user preferences for autonomous systems, but generally studies show that users of assistive devices prefer to maintain as much control authority as possible~\cite{argall_keynote,RSS2011_assistedteleoperation,lankenau2001}, and engaging the user is critical to robotic training~\cite{marchal2009}. Overconstraining user inputs may prevent them from utilizing certain valid control strategies. For instance, strict obstacle avoidance controls prevent wheelchair users from making maneuvers that bring them too close to a wall~\cite{lankenau2001}. Users may be willing to accept loss of control authority, but only if the improvements in performance are significant~\cite{RSS2011_assistedteleoperation, lankenau2001}. Therefore, devices are more likely to be used if they make tasks significantly easier without limiting users' actions~\cite{survey_deviceabandonment}. 

In robotic training, providing too much assistance or overconstraining users undermines the therapeutic impact of the device. Therefore many shared control schemes adapt their level of support online~\cite{emken2008,riener2005,wolbrecht2008, ellis2009} using an algorithm or schedule that prescribes changes based on some notion of the user's need for assistance. These levels of support can be modulated based on performance measures such as error~\cite{fisher2014,marchal2009, reinkensmeyer2016intro,patton2006error}, movement speed~\cite{kahn2004}, and task adeptness~\cite{krebs2003}, or based on the user's strength and fitness level~\cite{lokomat_clinical_study,RSS_exo2013} or current cognitive engagement in the task \cite{assistance_in_distraction}. At other times, the level of support can be manually adjusted by a physical therapist or the users themselves \cite{exoskeletons}. 

User trust in the robot is another critical factor in the overall performance of the joint human-machine system \cite{measurement_of_trust}. Trust, in this context, mainly depends on robot performance and its attributes, such as dependability, predictability, and level of automation \cite{trust_factors}. Thus, to encourage user trust, shared control algorithms should avoid robot behavior that is difficult for the human to understand~\cite{RSS_RoboObj2017}, unpredictable, or unnecessary. In task-based assistance, avoiding such behavior can be challenging, because there are often many ways of accomplishing a task and the individual is likely to take an approach that is different from the controller’s calculated strategy. Some shared control schemes have already been developed to adapt in real-time to user strategies~\cite{RSS_shah_adaptive}.

The primary contribution of this paper is a method for evaluating and selecting admissible user input. We present an assessment criterion that can be used in shared control schemes to improve training or performance while remaining minimally-engaging and flexible with respect to the user's approach. As our criterion for evaluation, we use the Mode Insertion Gradient (MIG)---a concept from hybrid control theory discussed in more detail in Section III. By calculating the MIG, one can assess how users' inputs affect the human-machine system over a time window into the future and allow inputs that are safe and/or do not hinder achieving a task goal.

Through a healthy human subject study, we show a correlation between user skill-level and the acceptance rate of the algorithm. Because we do not simply compare the user and controller decision at each time instance, we avoid the pitfall of arbitrarily rejecting actions that do not align with the controller's strategy but otherwise bring the system closer to a target configuration. In a sense, trust in the user is implicitly represented in the algorithm through the instantaneous assessment of user actions. Therefore, there is no need to implement an adaptive scheme that explicitly assesses user trustworthiness over time. Finally, in the human subject study ($n=28$), we observe that a MIG-based filter exhibits a training effect compared to training with no assistance for the tested group; in simulation, we demonstrate that a filtering algorithm utilizing a MIG criterion succeeds in task completion even with Gaussian noise inputs for two dynamic tasks---cart-pendulum inversion and SLIP balancing, while intervening minimally and remaining flexible with respect to the user's approach strategy.

\section{Methods}

\begin{figure}[h]
\centering
\includegraphics[width=0.9\columnwidth]{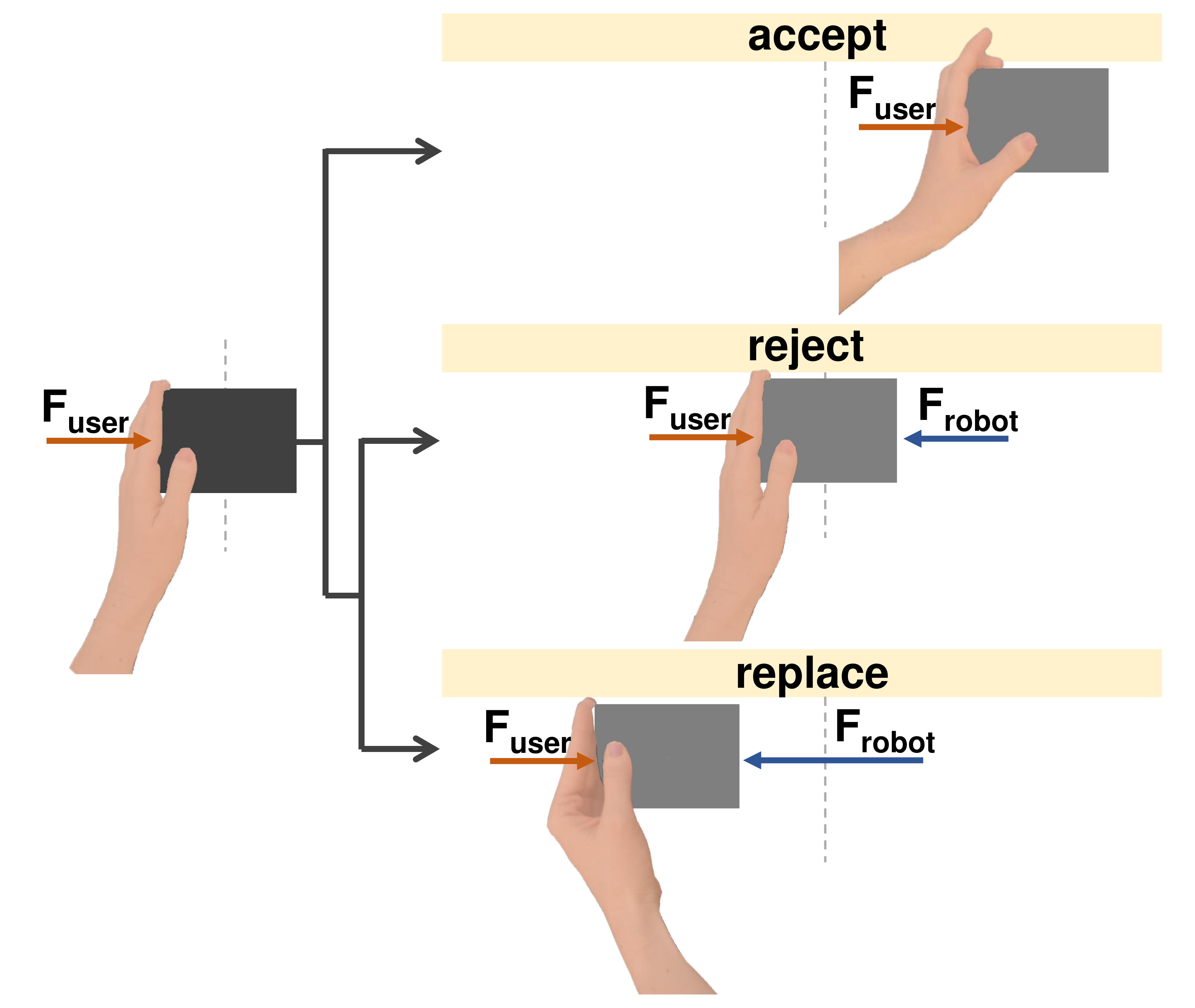}
\caption{Filter-based robotic responses on the example of a hand pushing a mass. The robot filters user input by physically accepting, rejecting, or replacing it. When a user action is accepted, the robot admits the force. When a user action is not accepted, the robot either rejects it by applying an equal and opposite force or replaces it by applying a force, such that the net effect on the system is equal to the controller-calculated input. }
\label{fig: accept_reject_replace}
\end{figure}

We conducted a human subject study, where we implemented and tested a shared control paradigm in the form of a mechanical filter (as shown in Fig.~\ref{fig: accept_reject_replace}). Subjects used an upper limb robotic platform as an interface to control a simulated cart-pendulum system with state vector $x=[\theta, \dot{\theta}, x_c, \dot{x_c}]$ and horizontal acceleration of the cart as control input. During experimental trials, the users' goal was to invert the pendulum to its unstable equilibrium. User input was inferred from a force sensor at the robot's end-effector and was continually evaluated at 100Hz. During trials when the filter was engaged, user actions were either accepted or rejected based on the criterion described in Section ~\ref{criterion}. 

\subsection{Experimental Platform}

\begin{figure}[t]
\centering
\includegraphics[width=1\columnwidth]{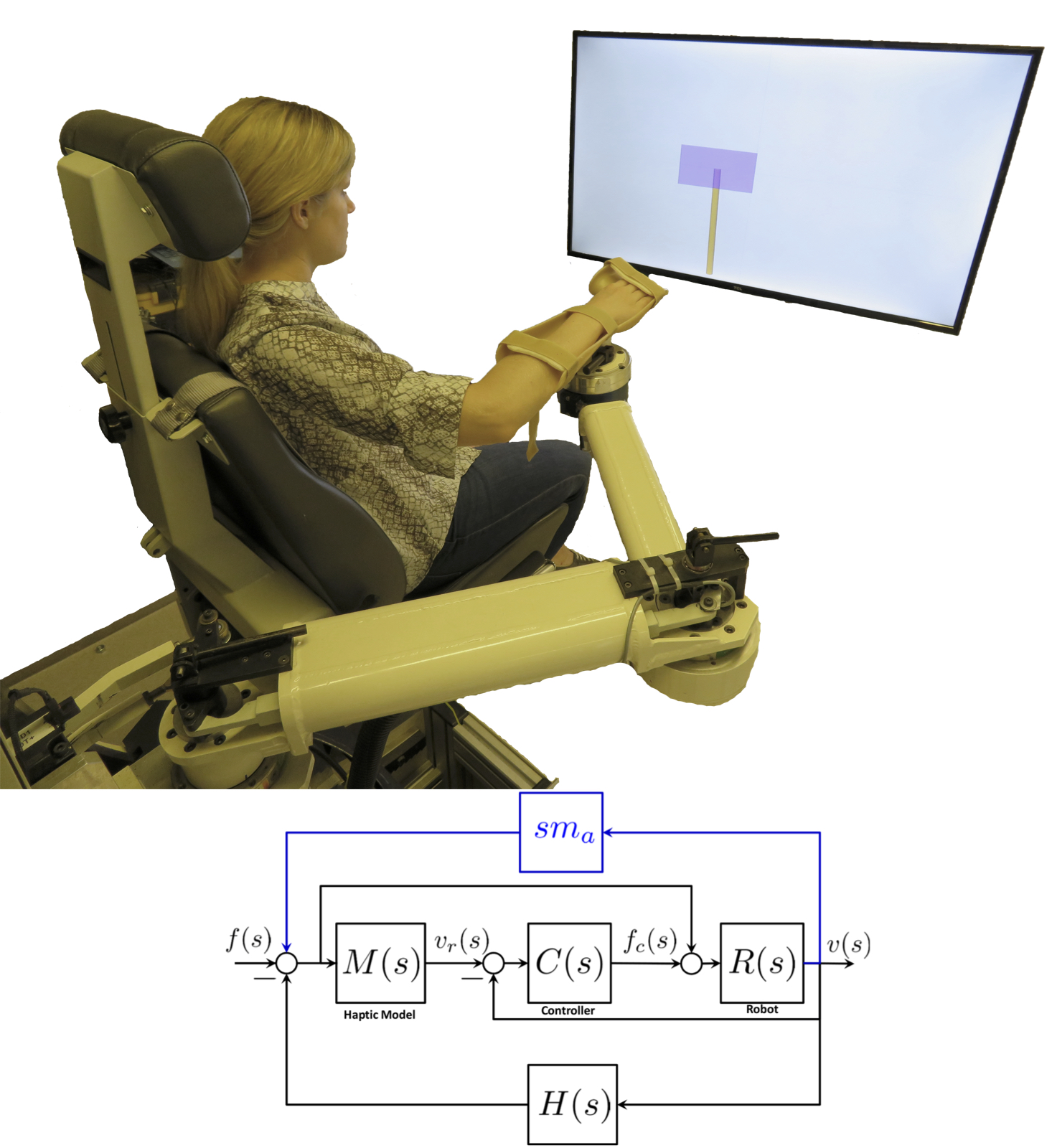}
\caption{(top) Upper limb robotic platform used during experiments. (bottom) The platform provides haptic feedback to simulate a specified inertial model via an admittance control scheme. A voluntary force $f(s)$ is measured by a force-torque sensor at the end-effector and passed through a model $M(s)$ that determines velocity $v_r(s)$ at which the robot should move. The reference velocity is tracked by the low level velocity controls, $C(s)$, of each motor drive. In addition to a force input, the user delivers involuntary impedance forces due to movement, given by dynamics $H(s)$. Acceleration information is fed back as a pseudo-force $sm_a$ for extra inertia reduction of the system.}
\label{nact3d}
\end{figure}

All human subject data was collected using the robotic platform shown in Fig.~\ref{nact3d}. The device is a powerful haptic admittance-controlled robot that can be used to render virtual objects, forces, or perturbations in three degrees of freedom. It is similar to the robotic platform used in \cite{ellis2016} and \cite{stienen2011} to provide a means to modulate limb weight support during reaching and to quantify upper limb motor impairments in stroke survivors.

During the experiment, the subject was seated in a Biodex chair with their arm secured in a forearm-wrist-hand orthosis. The orthosis could rotate passively, and the device could move its end-effector within a workspace defined both by its design limits and limits set by the investigators. At the point where the orthosis was mounted, a force-torque sensor measured subject input, which was then fed back to the admittance controller. In our experiments, the device was set up to physically support the upper limb of the participant in the z-direction while allowing them to move freely on the x-y plane. 

During testing, a display provided real-time visual state feedback to the user about the cart-pendulum system they were attempting to invert. High stiffness virtual springs in the haptic model were used to restrict user motion to a horizontal plane corresponding to the path of the cart in the virtual display. When user inputs were accepted, the control scheme behaved as described in Fig.~\ref{nact3d} and the end-effector motion changed according to the applied force. When user inputs were rejected, the measured user input $f(s)$ was ignored in the control scheme, such that the robot continued to move under its predefined dynamics as if no force had been applied by the user.

\subsection{Experimental Protocol} 

Twenty-eight subjects (9 males and 19 females) consented to participate in this study.\footnote{This study protocol was approved by the Institutional Review Board and all participants signed an informed consent form.} All subjects completed three sets of thirty 30-second trials with short breaks between sets. Each trial consisted of the subject attempting to invert a simulated cart-pendulum system, using cart acceleration as input. At the beginning of each session, the system and task was demonstrated to the subject using a video of a sample task completion. Subjects were instructed to attempt to swing up the simulated pendulum to the upward unstable equilibrium and balance it there for as long as possible. Subjects were instructed to continue to try to do this until the 30 seconds were over even if they succeeded at balancing near the equilibrium at some point throughout the trial.

Upon enrollment, subjects were randomly placed into either a control ($n=10$) or training group ($n=18$). During the second set, feedback in the form of a filter was engaged for the training group, while the control group completed each of the three sets without any feedback. Again, each user did three sets of thirty trials: set 1 (both groups: no feedback), set 2 (control: no feedback, training: feedback in the form of a mechanical filter), set 3 (both groups: no feedback).

\subsection{Performance Measures}\label{metrics}

Several measures were calculated to quantify user performance in individual trials. Specifically, time to success, balance time, and error were calculated for all trials and subsequently each trial was classified as successful or unsuccessful. 

A trial was considered successful when a subject reached an angle of $\pm0.4$~rad and angular velocity of $\pm0.75$~rad/s for at least 2 seconds. This success definition was used to determine the success rate and time to success of the users in each set. In addition, if a subject was successful, the total time spent at an angle of $\pm0.4$~rad and angular velocity of $\pm0.75$~rad/s 
\begin{figure}[!h]
\begin{center}
  	\includegraphics[width=1\columnwidth, keepaspectratio]{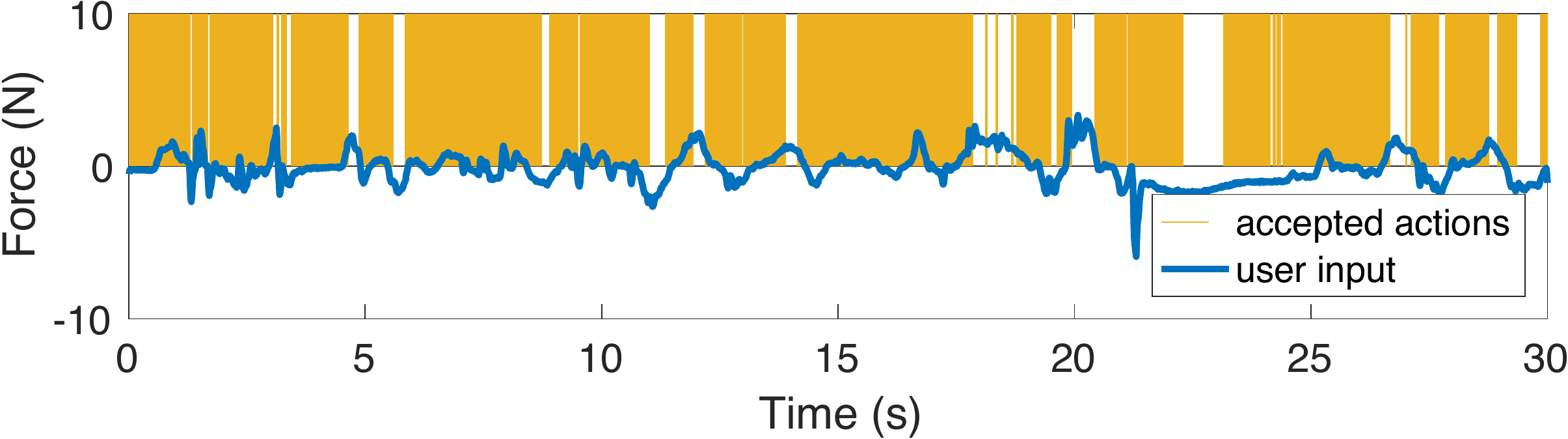}
  	\includegraphics[width=1\columnwidth, keepaspectratio]{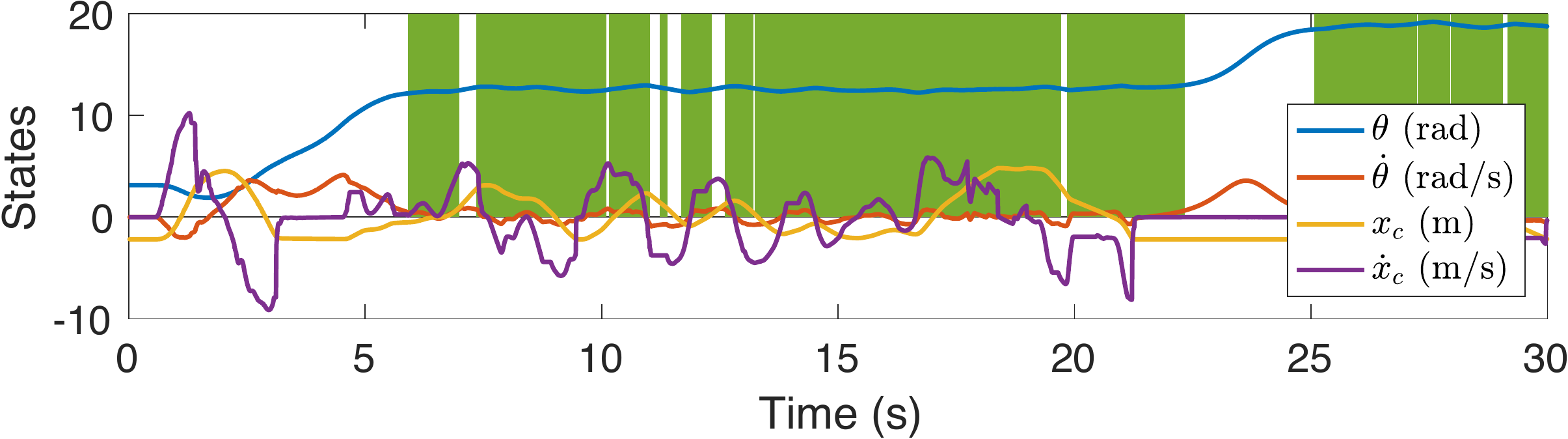}
\end{center}
\caption{Example trial data from study. (top) User force input with an indication of allowed actions in yellow. (bottom) System evolution with green highlighting of the time during which success was recorded. Note: Angle wrapping was not used on $\theta$ in the plot above, but it was used in the calculation of all performance measures. }
\label{fig: example_user}
\end{figure}
was recorded as the balance time. Note that when users were successful multiple times in the same trial, time spent in the balance region was cumulative. Lastly, an RMS error of each trajectory generated by the users was calculated with respect to the desired position in an inverted unstable equilibrium (zero-vector of the states). RMS error was normalized by the RMS error of a constant trajectory at the stable equilibrium, equivalent to the error of the user not moving from the initial conditions. 

A percent of rejected actions (PRA) was also recorded. PRA measured the fraction of user inputs that were rejected up to the time of a successful inversion, where we define an action to be a non-zero user input. 

Data from an example trial is visible in Fig.~\ref{fig: example_user}. In this case the trial was successful, with time to success = 8.3s, balance time = 19.7s, and RMS error = 0.57. The PRA was $13\%$.

\section{The Evaluation Criterion}

We present a new application for the Mode Insertion Gradient (MIG), which, to the authors' best knowledge, has not been previously used to assess human actions. Primarily a tool used in hybrid control theory, MIG can be interpreted as the sensitivity of a cost function to a discrete control input. Here, we use MIG to assess the impact of a user action on the evolution of a dynamic system over a time window into the future. We then utilize it as an evaluation criterion for a filter-based shared control paradigm and gather data to determine whether it serves as an objective, strategy-independent assessment criterion of user actions.

\subsection{Mode Insertion Gradient (MIG)}\label{criterion}

Usually, the mode insertion gradient $ \frac{dJ}{d\lambda}$ is used, in mode scheduling problems, to determine the optimal time $\tau$ to insert control modes from a predetermined set ~\cite{egerstedt2006,modesched2012,vaseduvan_switchsys2010, alex_SAC, caldwell2016}. Here we use the mode insertion gradient,
\begin{equation}\label{mig}
    \frac{dJ}{d\lambda}(\tau)=\rho(\tau)^T \left[f(x(\tau),u_2(\tau))-f(x(\tau),u_1(\tau))\right],
\end{equation}
as a measure of the sensitivity of the cost to a change from the nominal control, $u_1$, to a particular user input, $u_2$. In (\ref{mig}), state $x$ is calculated using nominal control, $u_1$, and $\rho$ is the adjoint variable calculated from the nominal trajectory,
\begin{equation*}
    \dot \rho = -\nabla l_1(x)-D_xf(x,u_1)^T\rho,
\end{equation*}
where $l_1(x,t)$ is the incremental cost and $\rho(t_f)=\nabla m(x(t_f))$. Moreover, in the work presented here, we assume the nominal control, $u_1$, to be equivalent to a null action ($u_1(t) = 0$), and we define $u_2$ with the piece-wise function below,
\[ u_2(t) =  \left\{
		\begin{array}{ll}
  			u_{user} & t\leq t_s \\
  			u_1 & t_s\leq t\leq T \\
		\end{array} 
\right. \]
where $t_s$ is the sampling time, $T$ is the time window over which we're evaluating system behavior, and $u_{user}$ is a user input recorded at current time $t$. It's worth noting that, in future work, $u_2$ could instead be defined by a combination of user input at current time $t$ and actions from an optimal controller over time $T$ into the future. This would add further flexibility to the criterion and give the user more control authority over the joint system, because any user action that could be corrected for by an optimal action without destabilizing the system during the time window $T$ would be admitted.  

When using MIG as an evaluation criterion, we calculate the integral of the mode insertion gradient over a time window $T$ into the future
\begin{equation}\label{mig_int}
    \int_{t_{now}}^{t_{now}+T}\frac{dJ}{d\lambda}(t)\delta t,
\end{equation}
to evaluate the impact of user control $u_2$ on the system over time $T$. When negative, the integral has been shown to indicate that $u_2$ is a descent direction over the entire time horizon \cite{Giorgos_NVC}, in a manner similar to the conjugate gradient descent method \cite{conjugate_gradient}, and thus serves as the basis for evaluating the impact of a current user action on the evolution of a dynamic system over that time window into the future. Moreover, stability can be inferred if (\ref{mig_int}) satisfies a contractive constraint \cite{contractive}. 

\noindent\begin{minipage}{\columnwidth}
\renewcommand\footnoterule{} 
\begin{algorithm}[H]
	\caption{A filter with MIG criterion.}\label{algo}
	\noindent\footnotetext{\noindent\normalsize *Note that the filter can be used with any model predictive controller (MPC) that can complete the task. Here a controller similar to \cite{alex_SAC} was used, but rather than using a single control value at a particular time as the control update, the entire control schedule was employed \cite{Giorgos_NVC, conjugate_gradient}.}
		    \\         
        Set sampling time $t_s$ and time horizon $T$. Set mode $m$ to either training $(T)$ or assistance $(A)$. Define objective function for filter and controller.
		\begin{algorithmic}[1]
			\While {$t_0 \leq t_f$}
			\State Infer user control vector $u_{user}$ from sensor data
			\State Simulate $x(t)$ and $\rho(t)$ in $[t_{now}, t_{now} + T]$ assuming \[ u =  \left\{
			\begin{array}{ll}
      			u_{user} & t\leq t_s \\
      			0 & t_s\leq t\leq T \\
			\end{array} 
			\right. \]
			\State Compute $\int \frac{dJ}{d\lambda}$
			\If {$ \int \frac{dJ}{d\lambda} \leq 0$}
			\State $u_{now} = u_{user}$
			\Else
			\If {$m=T$}
			\State Assign controller value
			$ u_{now} = 0 $
			\ElsIf {$m=A$}
			\State Calculate optimal control $u_{controller}$*
			\State Assign controller value 
			$ u_{now} = u_{controller} $
			\EndIf
			\EndIf
			\State Apply $u_{now}$ for $ t \in [t_{now}, t_{now}+t_s]$
			\EndWhile
		\end{algorithmic}	
	\end{algorithm}
	\renewcommand\footnoterule{}
\end{minipage}
\vspace{0.7cm}

In our experimental study, we utilize the MIG  criterion in a filter-based shared control scheme. For an outline of the approach, refer to Algorithm \ref{algo}. There are two modes for the MIG-based filter: a training and an assistance mode. In the training mode, no action is applied when the user's input is rejected. In the assistance mode, the robot is engaged to apply a controller-calculated action. An objective function defined as 
\begin{equation}
J = \frac{1}{2} \int_{t_o}^{t_f} \lVert x(t) - x_d(t)\rVert_{Q}^2 + \lVert u(t) \rVert_{R}^2 \delta t,
\end{equation}
with $Q \ge 0$ and $R \ge 0$ being metrics on state error and control effort and $x_d(t)$ being the desired trajectory, is used for the filter and model predictive controller (MPC).

\subsection{Simulated User Results}

In simulation, we show how controller intervention changes according to the skill level of a user. We note close to $0\%$ intervention for a simulated skilled user and $\sim50\%$ intervention for noise input in the one-dimension-controlled task of inverting a pendulum. 

\begin{figure}[h]
	\begin{center}
		\includegraphics[width=0.63\columnwidth, keepaspectratio]{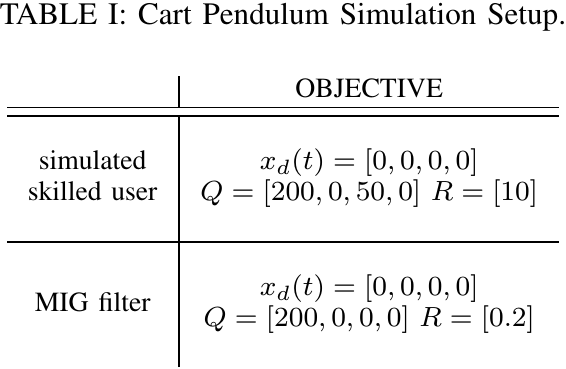}\par
	\end{center}
\end{figure}

To create the simulated skilled users, we utilize a model predictive controller with objectives representing successful inversion strategies. An example objective includes inverting the cart-pendulum while minimizing energy and staying close to the origin---the exact function parameters are given in Table~I. To approximate an unskilled user, we generate noise input. We then filter user actions using a MIG-based algorithm with a high-level objective function also listed in Table~I. There are many reasonable choices for the cost on the simulated users, but for the MIG filter, we chose to emphasize the goal of inversion by placing a high weight on the angle $\theta$.

Note that for simulated users the relationship between skill and controller intervention is explicit ($0\%$ intervention for an always successful user and $\sim50\%$ for noise input). With human subjects, we can only approximate their skill level and thus the relationship is more difficult to assess. 

\subsection{Human Study Results}

\begin{figure}[!t]
	\begin{center}
		\includegraphics[width=1\columnwidth, keepaspectratio]{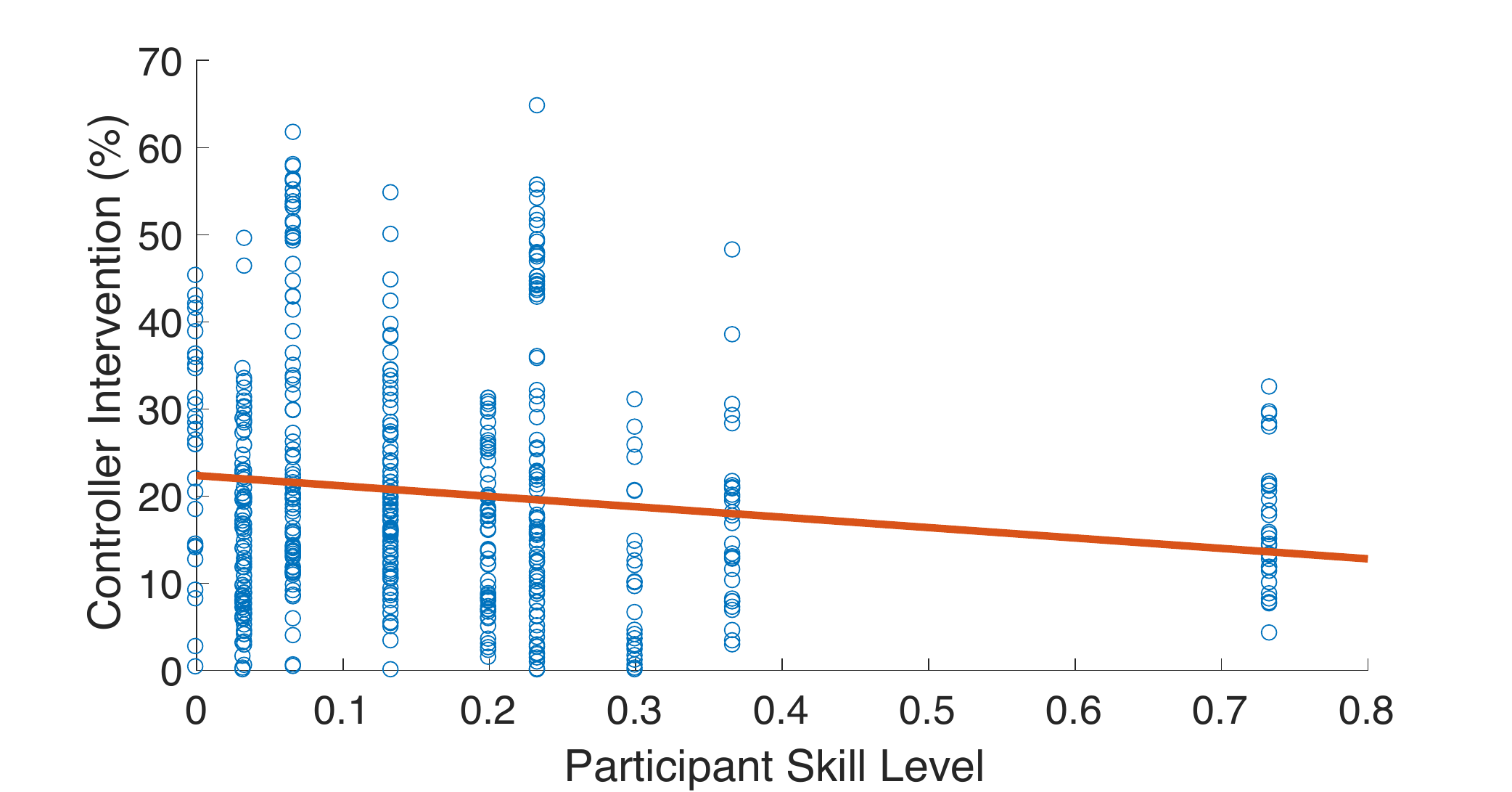}\par
	\end{center}
	\caption{A correlation coefficient of $-0.14$ is observed between the success rate of the users in set 1 with no assistance and the rejection rate of the users' inputs in set 2 with assistance, suggesting a correlation between the users' adeptness at the task and the controller's intervention rate during assistance. }
	\label{fig: MDA_corr_dotproduct}
\end{figure}

A human study was conducted to determine whether a relationship could be observed between participant skill level---estimated based on performance in unassisted trials---and the frequency of controller intervention in the MIG filter mode. In this case, we calculate the success rate of the 30 trials from set 1 to approximate user skill level. We then use percent of rejected actions (PRA) values from individual trials in set 2 from the same users to identify the correlation---a scatter plot with the results is visible in Fig.~\ref{fig: MDA_corr_dotproduct}. A Pearson product-moment correlation coefficient was computed and a low negative correlation ($r=-0.14$, confidence interval $(-0.22)-(-0.06)$, $p=0.001$) was identified between overall success rate in set 1 and PRA in individual trials of set 2 for the training group ($n=18$). Similar but weaker correlations were identified between controller intervention and other task-specific metrics, such as balance time ($r=-0.09$, confidence interval $(-0.17)-(-0.007 )$, $p=0.03$) and time to success ($r=0.11$, confidence interval $0.086-0.25$, $p=0.01$).

Since subjects showed significant improvement during set 1 while getting used to the task and testing platform, we ran the same statistics using only the last 10 trials of set 1 to estimate participant skill level. Again, a Pearson product-moment correlation coefficient was computed and a low negative correlation ($r=-0.20$, confidence interval $(-0.28)-(-0.11)$, $p=4.5 \cdot 10^{-6}$) was identified between overall success rate in the last 10 trails of set 1 and PRA in individual trials of set 2. Similar correlations were identified between controller intervention and other task-specific metrics, such as balance time ($r=-0.13$, confidence interval $(-0.21)-(-0.04)$, $p=0.003$) and time to success ($r=0.21$, confidence interval $0.13-0.29$, $p=7.3 \cdot 10^{-7}$).

Overall, for an experimental group of 18 participants, we obtained low but significant correlations \cite{cohen_statistics} between independently measured performance metrics and rejection rate in assisted trials, suggesting a relationship between the users' skill level and the MIG filter's rate of intervention, respectively. Because the correlations are weak, additional subjects and analysis of other tasks are needed before the skill sensitivity is conclusive. However, our initial findings suggest that a MIG criterion is a skill-sensitive paradigm that can be used for shared control. As the next two sections detail, it substantially increases improvement during training as compared to training with no feedback and, in simulation, it improves task success and safety when used for assistance.

\section{MIG for Training}

\begin{figure}[t]
	\begin{center}
		\includegraphics[width=\columnwidth]{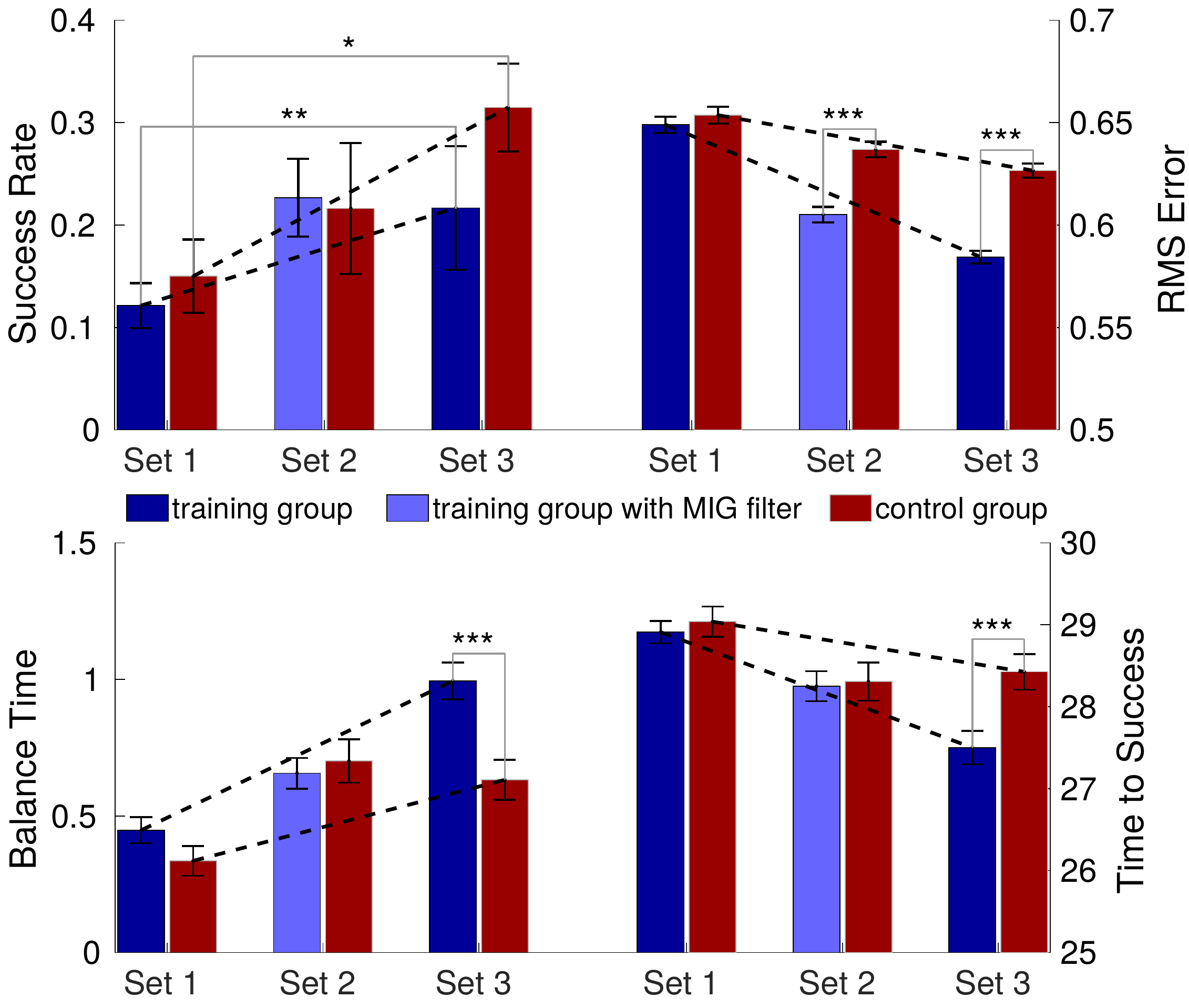}
	\end{center}
	\caption{Set was consistently the most significant factor in performance improvements from set 1 to set 3. As expected, pairwise comparisons of the two groups in set 1 show that there was not a significant difference in their baseline performance measurements. However, the RMS error, balance time, and time to success of the training group in the final set was significantly better than that of the control group. Note: error bars indicate standard error; significance is indicated by $^*p<0.05$, $^{**}p<0.01$, $^{***}p<0.001$.}
	\label{fig: training}
\end{figure}

A two-factor repeated measures ANOVA was used to assess the effects of the group (between-subjects) and set (within-subjects) on all performance measures listed in section~\ref{metrics} (Fig.~\ref{fig: training}). The training group and control group were evaluated based on set 1 and set 3 only. Set 2 was left out of the ANOVA, so that effects of the assistance itself would not be measured in the analysis.

The factorial ANOVA revealed that the effect of group ($F(1,50)=0.981,\ p=0.327$) and the interaction effect ($F(1,50)=0.111,\ p=0.740$) of the group  and set on the success rate were not significant. The main effect of set yielded an F ratio of $F(1,50)=7.555,\ p=0.008$, meaning that users were more successful in set 3 ($mean =0.280 , SD = 0.223$) than in set 1 ($mean = 0.140,\ SD = 0.100$) regardless of the type of practice in set 2. Pairwise comparisons were made between set 1 and set 3  of each group using a paired 2 sample t-test. The change in success rate from set 1 to set 3 was significant for the control group ($t(9)=3.152,\ p = 0.012$) and the experimental group ($t(17)=3.127,\ p=0.006$). Although set was the predominant factor in success rate, the the effect size of the training group ($d=0.94$)  from set 1 to set 3 was larger than the effect size of the control group ($d=0.77$). Note that the control group continued to improve their success rate with each set, possibly because their interaction with the robot did not change between sets as it did for the training group. Pairwise t-tests of the training group and control group showed that the difference in success rate between the training group and the control group was not significant for any of the three sets.

A factorial analysis of variance, evaluating the impact of training group and set on the RMS error showed that the main effect of group ($F(1,26)=1.615, p = 0.215$) was not significant. Therefore, there was not a significant difference between the training group ($mean =0.612, SD =0.088$) and the control group ($mean = 0.639 , SD=0.067$). The main effect of set yielded an F ratio of $F(1,26) = 41.551, p<0.001$ indicating a significant difference between set 1 ($mean =0.651, SD =0.085$) and set 3 ($mean =0.599, SD =0.072$). The interaction effect of group and set was significant ($F(1,26)=5.099, p=0.0326$), implying that the training had a greater impact on set 3 performance than the unassisted practice of the control group. 

There was no significant effect of group on balance time ($F(1,26)=1.562,\ p=0.223$) or time to success ($F(1,26) = 1.114,\ p=0.301$). There was also no significant interaction effect of group and set on balance time ($F(1,26)=1.048,\ p=0.315$) or time to success ($F(1,26)=1.512,\ p=0.22983$). The main effect for set on the balance time yielded an F ratio of  $F(1,26)=15.328,\ p<0.001$, indicating a significant difference between the balance time in set 1 ($mean = 0.408, SD = 1.053$) and set 3 ($mean = 0.866, SD = 1.476$). The main effect of set on time to success was also significant ($F(1,26)=18.992,\ p<0.001$), with set 3 ($mean = 27.830, SD = 4.433$) outperforming set 1 ($mean = 28.955,SD =3.175$). According to 2-sample t-tests, the difference between the balance time of the control group and training group in set 1 was not significant, but the set 3 balance time of the control group ($mean =0.632, SD = 1.261$) was significantly less ($t(728)=3.643,p<0.001$) than the balance time of the training group ($mean = 0.994 , SD = 1.568$). The time to success was also significantly better ($t(738)=3.110, p=0.002$) in set 3 of the training group ($mean = 27.500, SD = 4.744$) compared to set 3 of the control group ($mean =28.43, SD = 3.74$).

In summary, pairwise comparisons within each of the four measures (success rate, RMS error, balance time, and time to success) showed that in set 1 there was not a significant difference between the training group and control group, suggesting that on average the two groups started off with the same skill at the task. Set had a significant effect on increases across all metrics, indicating that participants were continuously improving with time regardless of the feedback that was provided. Although there was not a significant effect of group on any of the metrics, the RMS error showed that there was a significant interaction effect between group and set. This is indicated in Fig.~\ref{fig: training} by the two groups having similar means in set 1 but significantly different means in set 3. Moreover, when training group and control group were compared in set 3, the training group performed significantly better. Finally, we observe that when in use during set 2 of the training group, the MIG filter had a significant effect on reducing the RMS error, while it did not have a significant effect on success rate, balance time or time to success. Hence, we can reason that the MIG filter guided users through the task without getting in the way or accomplishing the task for them.

\section{MIG for Assistance}

Whereas during training, we allow users to fail at task completion for improved learning, during assistance in tasks, such as activities of daily living (ADL), we may want to insist on task success, user safety, or both. In these situations, we can modify the proposed filter to actively provide assistance. Instead of using a null controller input as the alternative to user input, we can engage the controller and replace rejected actions with optimal control, calculated by an MPC. In the next two subsections, we provide simulation results that demonstrate system behavior when the MIG-based filter is employed in assistance mode.

\subsection{Cart Pendulum - Task Completion}

\begin{figure}[!t]
\begin{center}
    \includegraphics[width=1\columnwidth, keepaspectratio]{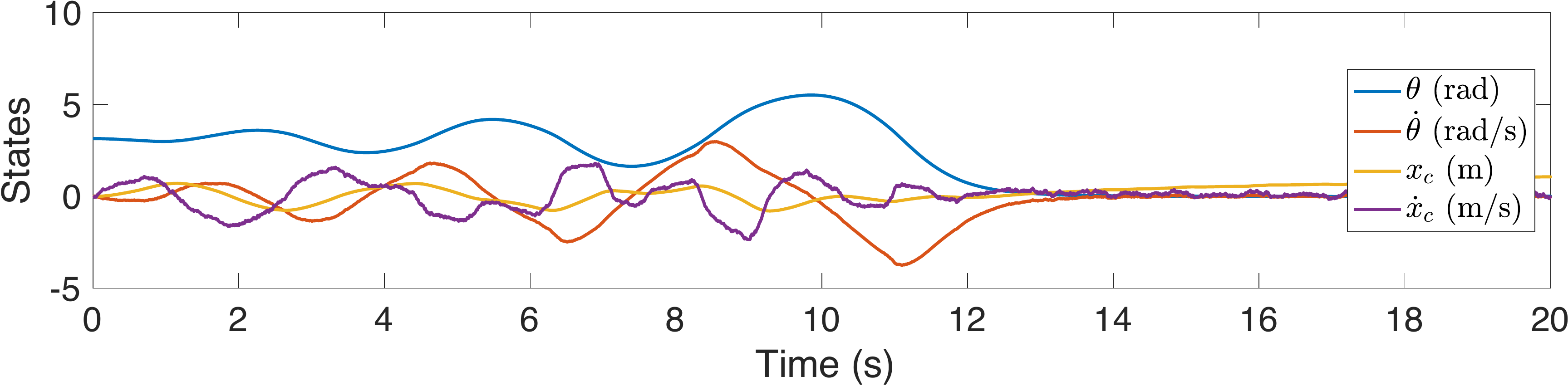}
    \includegraphics[width=1\columnwidth, keepaspectratio]{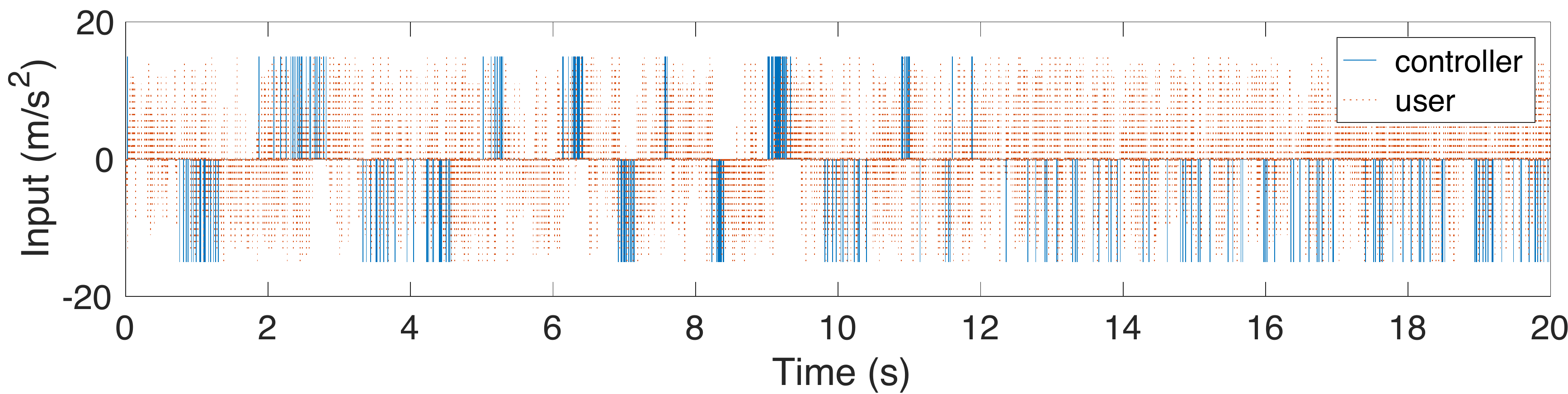}
    \includegraphics[width=1\columnwidth, keepaspectratio]{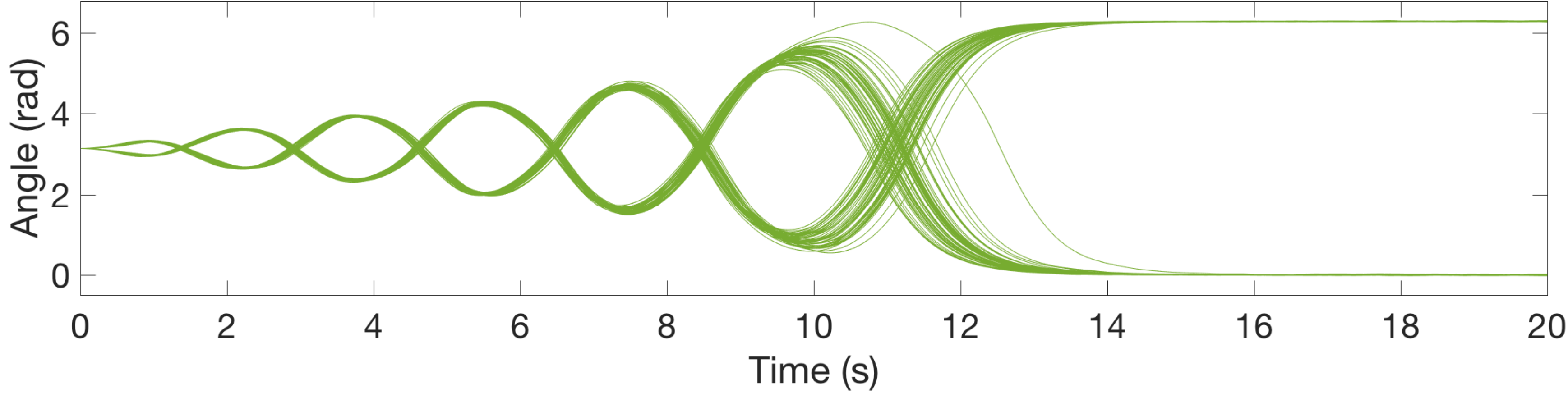}
\end{center}
\caption{For the cart pendulum inversion task, noise input with a MIG-based filter in assistance mode is able to invert the pendulum in 100 out of 100 of the simulation ran. (top, middle) An example trial with the system evolution and filtered input are shown. (bottom) Convergence results from all 100 trials.}
\label{fig: MC_cp}
\end{figure}

A series of 100 Monte Carlo simulations demonstrate a $100\%$ success rate for filtered noise input in the cart pendulum inversion task, suggesting that a MIG-based filter could be employed in situations where task completion is crucial. System behavior, simulated user input, and controller intervention during an example trial are visible in Fig.~\ref{fig: MC_cp}. Results of the 100 trials with noise input are also shown.

\subsection{SLIP - Safety}

Lastly, we analyze the performance of MIG-based assistance on a spring-loaded inverted pendulum (SLIP) model. The SLIP is a hybrid, low-dimensional system that has been shown to be a reliable approximation of human running \cite{SLIP_for_running_nature} and is therefore used to model running dynamics in robotic locomotion \cite{SLIP_robot_evidence}. Here, a 2D SLIP model is tested with a state vector described by $x = [x_m, \dot{x}_m, z_m, \dot{z}_m, x_t]$, where $x_m$ and $z_m$ are the coordinates of the mass, and $x_t$ is the coordinate of the toe, and a control vector described by $u = [u_s,u_t]$, where $u_s$ is the leg thrust applied during stance and $u_t$ is the toe velocity control applied during flight. Hybrid dynamics of the form
\begin{align*}
    f_{stance}= 
    \begin{pmatrix}
    \dot{x}_m \\
    \frac{(k(l_0-l_s)+u_s)(x_m-x_t)}{ml_s} \\
    \dot{z}_m \\
    \frac{(k(l_0-l_s)+u_s)z_m}{ml_s} - g \\
    0
    \end{pmatrix}
\end{align*}
and $f_{flight} = (\dot{x}_m,0,\dot{z}_m,-g,\dot{x}_m+u_t) $
are used. Parameters $k$, $l_0$, and $m$ describe the SLIP model spring constant, resting spring length, and mass, respectively. All parameters were given a value of 1 in our simulations. To determine switches between stance and flight modes, a guard equation $\phi(x)$ is employed
\begin{equation*}
    \phi_{stance \rightarrow flight}(x) = \phi_{flight \rightarrow stance}(x) = x_m - \frac{l_0}{l_s}z_m
\end{equation*}
with $l_s$ being the leg length during stance
\begin{equation*}
    l_s=\sqrt{(x_m-x_t)^2+z_m^2}.
\end{equation*}

In the experiments, we use input from simulated users of different skill level, which we generate using MPC with objective functions outlined in Table~II. We approximate an unskilled user using Gaussian noise; a low-skill user using MPC with a height objective lower than the spring length that causes the SLIP to fall; and a skilled user using MPC with a feasible objective such that the controller can achieve forward motion without assistance. 

\begin{figure}[!t]
\begin{center}
   	\includegraphics[width=1\columnwidth, keepaspectratio]{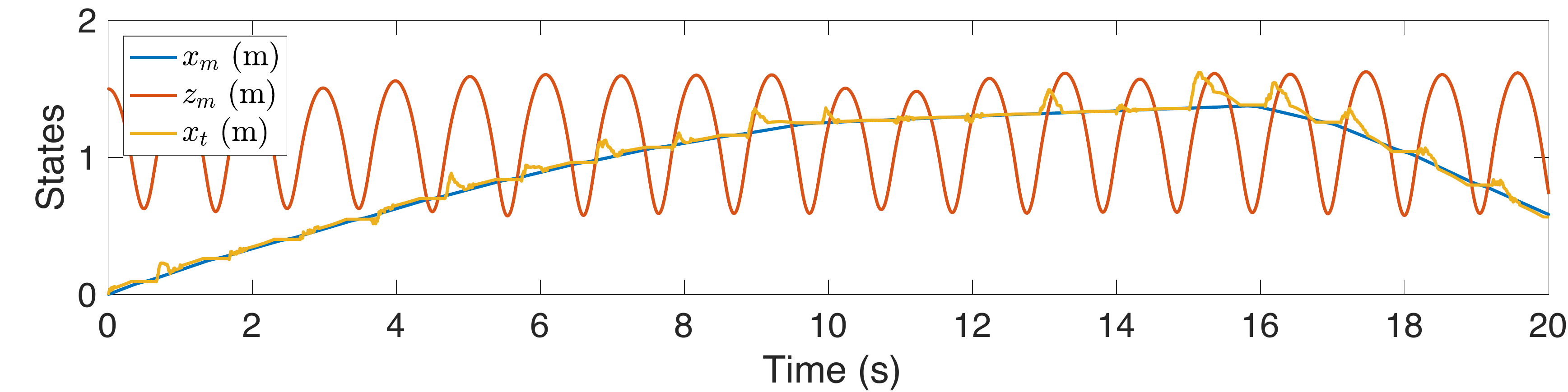}
\end{center}
\caption{We simulate a SLIP model using Gaussian noise as user input and the MIG-filter in assistance mode for support. Note that the filter allows the foot to make random movements and the SLIP to change direction, while keeping the center of mass oscillating around a safe constant height. The controller overrides the user's input for $\sim70\%$ of the simulation time.}

\label{fig: SLIP_noise}
\end{figure}

We show that with the MIG filter in assistance mode the SLIP can be kept upright even when input is provided by Gaussian noise or a low-skill user. From Fig.~\ref{fig: SLIP_noise} we see that for noise input the filter allows the foot to make random movements and the SLIP to change direction, while keeping the center of mass oscillating around a safe constant height. 

\begin{figure}[b!]
	\begin{center}
		\includegraphics[width=0.95\columnwidth, keepaspectratio]{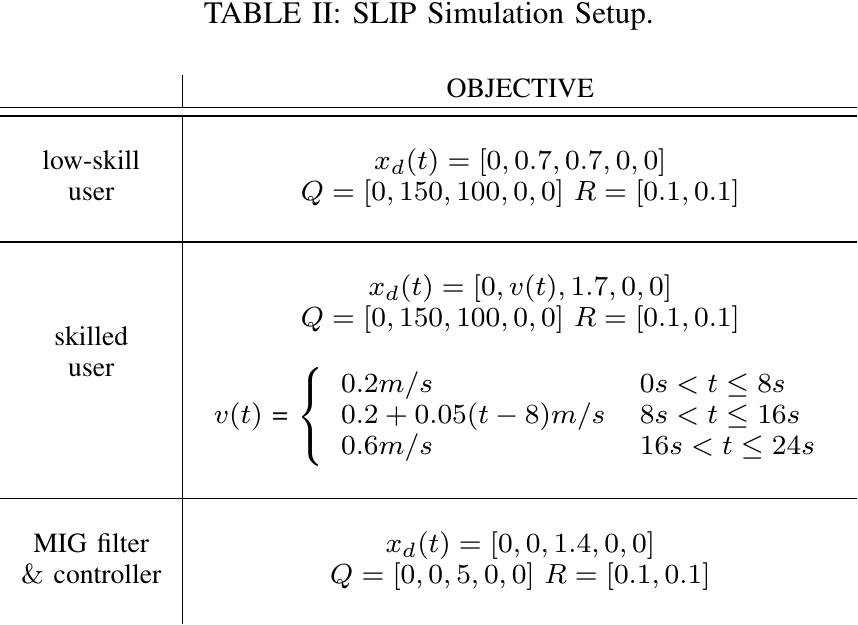}\par
	\end{center}
\end{figure}

\begin{figure}[!t]
\begin{center}
  	\includegraphics[width=1\columnwidth, keepaspectratio]{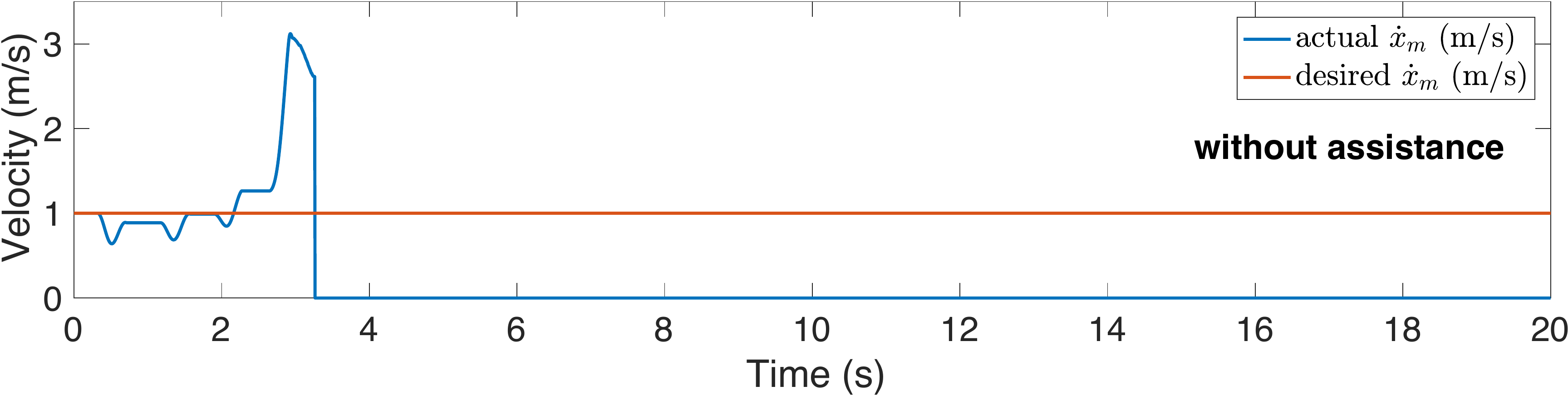}
  	\includegraphics[width=1\columnwidth, keepaspectratio]{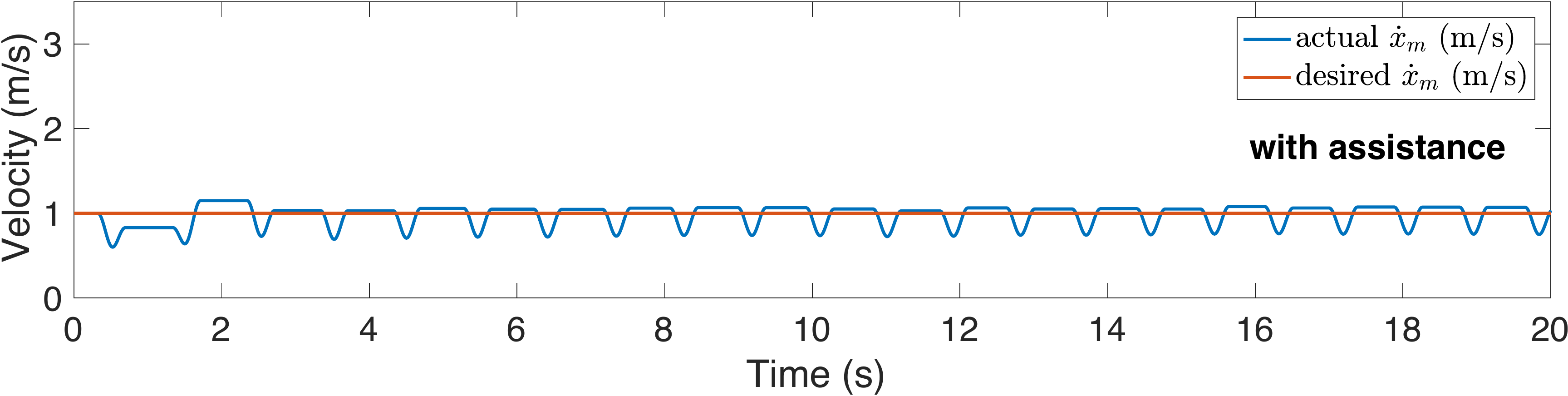}
\end{center}
\caption{(top) We simulate a low-skill user that attempts to move forward with no assistance. The SLIP falls after $\sim5.5s$. (bottom) We use the same user simulation but now the controller helps the user keep balance without restricting its forward motion. With under $40\%$ controller intervention, the SLIP establishes a cyclic gait and maintains an average speed of $0.98~m/s$ (close to the user's desired $1~m/s$).}
\label{fig: SLIP_weak}
\end{figure}

\begin{figure}[!t]
\begin{center}
   	\includegraphics[width=1\columnwidth, keepaspectratio]{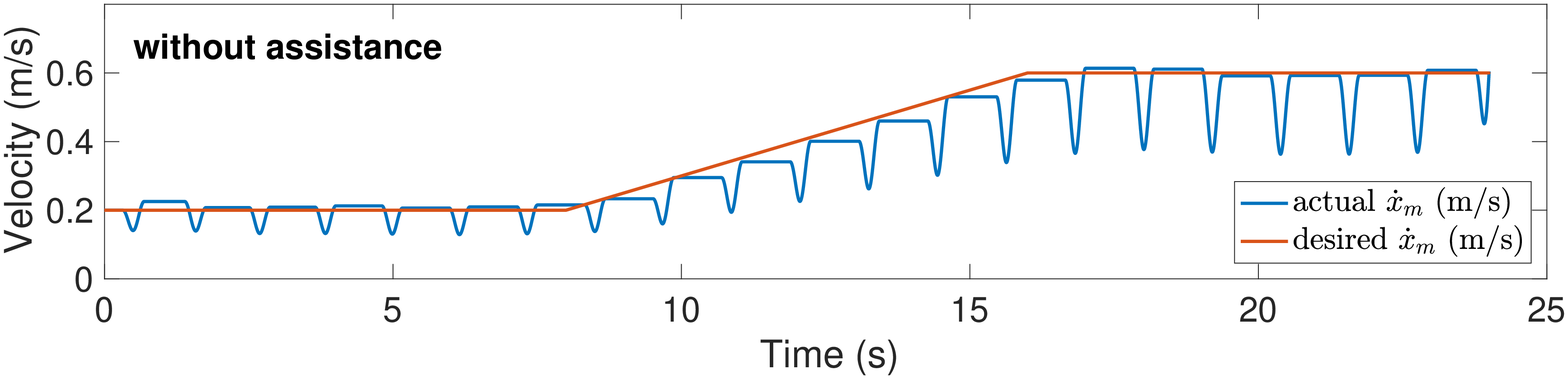}
   	\includegraphics[width=1\columnwidth, keepaspectratio]{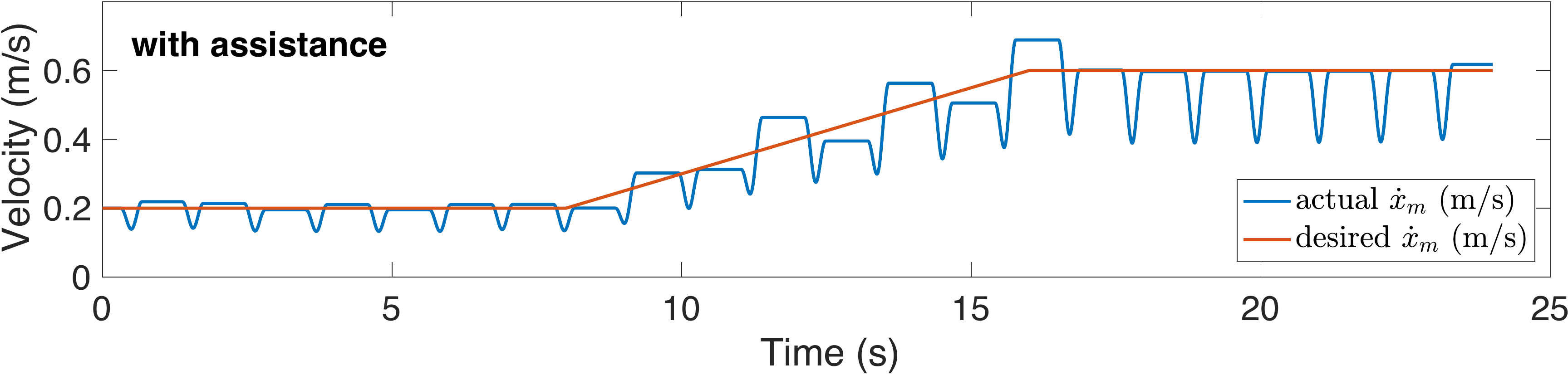}
\end{center}
\caption{(top) We simulate a capable user that attempts to move forward with varying velocity. (bottom) We simulate the same user with added assistance. Note that the assistance does not impede the user's forward motion, even though the controller has no \textit{a priori} knowledge of the user's desired velocity. The controller intervenes $\sim~20\%$ of the time.}
\label{fig: SLIP_skilled}
\end{figure}

For a low-skill user, the assistance prevents the SLIP from falling, while allowing it to maintain its desired forward velocity, as visible in Fig.~\ref{fig: SLIP_weak}. Finally, when provided with input from a skilled user, the filter allows the user to dictate its desired forward velocity and interferes only minimally with its desired motion, as visible in Fig.~\ref{fig: SLIP_skilled}. The controller overrides user input for $\sim70\%$ of the time for noise input, for under $40\%$ of the time for a low-skill user, and for $\sim~20\%$ of the time for a skilled user.

Based on these results, the MIG criterion shows promise to be used in applications, such as lower-limb exoskeletons \cite{exoskeletons}. In walking assistance, we want to at all cost prevent users from falling, while at the same time giving them freedom to follow their natural gait pattern, walk at a desired pace, and change speeds or stop when convenient. 

\section{Conclusion} 
\label{sec:conclusion}

A variety of shared control paradigms have been implemented to provide assistance to users in settings where the task is known \textit{a priori}. Although users might prefer to maintain control and user engagement is necessary for learning, many applications require a certain level of control authority to be allocated to the machine in order to guarantee safety, successful task completion, or both. As such, most interfaces employ support strategies that in various ways restrict or adjust users' actions in order to enable the subject and the device to move in a safe and stable manner. In this paper, we present and evaluate an assessment criterion for user input that can be utilized in these shared control paradigms. We carry out experiments by using the MIG as an evaluation criterion in a filtering assistance scheme, similarly to \cite{hapticsharedcontrol_review, MDA_Katie}, where user actions deemed by the filter as incorrect are either blocked or hindered by the hardware interface. 

With only current state information, our proposed filter can both reject unhelpful inputs and remain transparent to operators with significant skill. For complex dynamic tasks, such as walking with an exoskeleton, the algorithm can help provide meaningful assistance and ensure safety of the system and operator without limiting the user's freedom. It can, like adaptive methods, enhance human-system performance while avoiding some of the common long-term pitfalls of ``static" automation such as over-reliance, skill degradation, and reduced situation awareness \cite{adaptive_review}. 

\section*{Acknowledgments}

This work was supported by the National Science Foundation under grants 1329891 and 1637764 and by the National Defense Science and Engineering Graduate Fellowship program. Any opinions, findings, and conclusions or recommendations expressed in this material are those of the authors and do not necessarily reflect the views of the National Science Foundation or of the NDSEG program.

The authors would like to thank Sabeen Admani for her
unwavering support in debugging the robot and keeping our human experiments on schedule. 

\bibliographystyle{plainnat}
\bibliography{references}

\end{document}